\documentclass[conference,11pt]{IEEEtran}

\IEEEoverridecommandlockouts
\usepackage[T1]{fontenc}
\usepackage{newtxtext,newtxmath}
\usepackage{amsmath,amssymb,amsfonts}
\usepackage{algorithmic}
\usepackage{graphicx}
\usepackage{textcomp}
\usepackage{xcolor}
\usepackage{times}
\usepackage{epsfig}
\usepackage{kotex}
\usepackage{booktabs}
\usepackage[normalem]{ulem}
\usepackage{tikz}
\usepackage{cite}
\usepackage{caption}
\usepackage{subcaption}
\usepackage{array}
\usepackage{multirow}
\usepackage{hyphenat}
\usepackage{balance} 
\usepackage{tikz}


\def\BibTeX{{\rm B\kern-.05em{\sc i\kern-.025em b}\kern-.08em
 T\kern-.1667em\lower.7ex\hbox{E}\kern-.125emX}}

\begin{document}

\author{\IEEEauthorblockN{Rebira Jemama}
\IEEEauthorblockA{\textit{Lewisburg Area High School, Lewisburg, USA } \\
Rebirajemama01@gmail.com}
\and
\IEEEauthorblockN{Rajesh Kumar}
\IEEEauthorblockA{\textit{Bucknell University, Lewisburg, USA}\\
rk042@bucknell.edu}
}

\title{How Well Do LLMs Imitate Human Writing Style? \thanks{The paper is accepted for publication at IEEE UEMCON 2025. \\ More at https://github.com/rajeshjnu2006/writing-style-uemcon2025}}

\maketitle

\begin{abstract}
Large language models (LLMs) can generate fluent text, but their ability to replicate the distinctive style of a specific human author remains unclear. We present a fast, training-free framework for authorship verification and style imitation analysis. The method integrates TF-IDF character n-grams with transformer embeddings and classifies text pairs through empirical distance distributions, eliminating the need for supervised training or threshold tuning. It achieves $97.5\%$ accuracy on academic essays and $94.5\%$ in cross-domain evaluation, while reducing training time by $91.8\%$ and memory usage by $59\%$ relative to parameter-based baselines. Using this framework, we evaluate five LLMs from three separate families (Llama, Qwen, Mixtral) across four prompting strategies: zero-shot, one-shot, few-shot, and text completion. Results show that the prompting strategy has a more substantial influence on style fidelity than model size: few-shot prompting yields up to $23.5$x higher style-matching accuracy than zero-shot, and completion prompting reaches $99.9\%$ agreement with the original author’s style. Crucially, high-fidelity imitation does not imply human-like unpredictability: human essays average a perplexity of $29.5$, whereas matched LLM outputs average only $15.2$. These findings demonstrate that stylistic fidelity and statistical detectability are separable, establishing a reproducible basis for future work in authorship modeling, detection, and identity-conditioned generation.
\end{abstract}


\section{Introduction}  
\label{Introduction}
Large language models (LLMs) now generate text that is fluent, coherent, and adaptable across a wide range of domains~\cite{surveyLLMDetection}. Their ability to mimic style and tone creates both opportunities and risks. Personalized generation can enhance education, accessibility, and creativity; however, the same ability also threatens authorship integrity in scholarly work, fuels misinformation, and complicates forensic investigations. These tensions motivate a central question: \textit{can an LLM reproduce the stylistic fingerprint of a human author while remaining statistically detectable as machine-generated?}

Style imitation goes beyond topical accuracy. It involves recurring author-specific cues such as consistent sentence length, characteristic punctuation habits (for example, heavy use of semicolons), lexical preferences, or syntactic constructions. These signals form the basis of stylometry and authorship analysis, which have long been applied in forensics, plagiarism detection, and literary studies  \cite{stamatatos2009survey,sapkota2015not}. Parallel to this, a separate line of work addresses machine-generated text detection using perplexity~\cite{jelinek1977perplexity,radford2019gpt2}, fine-tuned classifiers, and watermarking methods~\cite{DetectGPT}. However, detectors often fail across domains, and humans themselves frequently misclassify LLM text as human-authored~\cite{surveyLLMDetection}.  

These two literatures—stylometry and text detection—have rarely been combined. Stylometry focuses on distinguishing among human authors, whereas detection targets the boundary between human and machine-generated texts. Recent work suggests that these perspectives can converge: LLM detection may be framed as an authorship verification problem rather than a pure attribution problem~\cite {authorshipVerificationLLM}. Yet, what is missing is a reproducible, model-agnostic framework that directly quantifies how closely LLM-generated text matches a target author’s style and relates this fidelity to detectability.  

This study aims to fill that gap. We pursue four research questions:  
\underline{$q_1$}: How can we design a reproducible and scalable protocol that quantifies stylistic similarity between human-authored and LLM-generated texts without subjective judgments or task-specific training?  
\underline{$q_2$}: How do different LLMs and prompting strategies compare in style imitation fidelity under matched experimental conditions?  
\underline{$q_3$}: Does high-fidelity imitation also yield human-like unpredictability, or do generated texts remain statistically identifiable as synthetic~\cite{DetectGPT,radford2019gpt2}?  
\underline{$q_4$}: What verifier design enables efficient, large-scale experimentation on commodity hardware while preserving accuracy across domains~\cite{bevendorff2020pan20}? We answer these questions with four contributions:  

\begin{itemize}

\item First, we present a reproducible framework for measuring LLM style imitation accuracy ($q_1$). The framework is training-free and model-agnostic, combining TF--IDF character n-grams with transformer embeddings~\cite{sentencetransformers,wang2020minilm}. It scales to thousands of text pairs, avoids threshold tuning, and situates comparisons within intra- and inter-author distributions.  

\item  Second, we conduct a comparative evaluation of LLMs and prompting strategies ($q_2$). Using the framework, we benchmark two Llama variants~\cite{llama3.3-hf,llama4-scout-hf}, two Qwen variants~\cite{bai2023qwentechnicalreport}, and one Mixtral model~\cite{jiang2024mixtralexperts} across zero-shot, one-shot, few-shot, and text-completion prompts. The results show that the prompting strategy has a greater impact on stylistic fidelity than model size, with few-shot and completion settings achieving nearly perfect matches.  

\item Third, we analyze AI detectability in high-fidelity imitation scenarios ($q_3$). Although LLM outputs can closely reproduce human style, they remain statistically more regular than human text. Across $1,000$ essays, human-authored text averaged a perplexity of $29.5$, compared to $15.1$ for LLM outputs, confirming that stylistic fidelity and statistical detectability are separable.  

\item  Finally, we introduce a lightweight, training-free authorship verifier ($q_4$). The verifier constructs in under five seconds, reduces memory use by about 60 percent compared to supervised pipelines, and integrates interpretable stylometric features with contextual embeddings, enabling efficient and scalable experimentation.  

\end{itemize}  

The rest of the paper is organized as follows. Section \S\ref{RelatedWork} related work; \S\ref{MaterialsMethods} describes data, features, the distribution-based verifier, metrics, and prompting protocols; \S\ref{Results} reports verification, imitation, completion, and detectability results; followed by discussion on the results in \S\ref{Discussion}, with \S\ref{Limitations} listing limitations and \S\ref{Conclusion} concluding the paper.

\section{Related Work}
\label{RelatedWork}
Authorship analysis has a long history in computational linguistics, traditionally framed as attribution (assigning text to one of several candidate authors) or verification (deciding whether two texts come from the same author)~\cite{juola2006authorship,stamatatos2009survey}. Early approaches relied on stylometric features such as character and word n-grams, function word frequencies, punctuation habits, and syntactic patterns~\cite{sapkota2015not,segarra2014authorship,schler2006effects,alonso2021writer}. These cues capture unconscious linguistic choices and have proven effective in forensics and plagiarism detection. More recent work has explored neural approaches, ranging from syntactic recurrent networks~\cite{jafariakinabad2019syntactic} to convolutional classifiers for script style~\cite{gamback2017using}, BERT-based fine-tuning~\cite{fabien2020bertaa}, attention-based similarity learning~\cite{ExplainableAV}, and vector-difference methods designed for open-world verification~\cite{Weerasinghe2021Feature}. Shared evaluation campaigns such as Plagiarism Analysis, Authorship Identification, and Near-Duplicate Detection (PAN) have further standardized authorship verification benchmarks and highlighted the continued utility of character n-grams in combination with deep contextual embeddings~\cite{bevendorff2020pan20}.

In parallel, a growing literature examines the detection of machine-generated text. Early efforts focused on statistical metrics, such as perplexity~\cite{jelinek1977perplexity,radford2019gpt2}, with later work introducing curvature-based detectors, including DetectGPT~\cite{DetectGPT}. Empirical studies confirm that perplexity remains a strong signal, with human text averaging substantially higher unpredictability than model outputs~\cite{Gutierrez-Megias2024perplexity}. A recent survey consolidates these developments, emphasizing both the necessity and limitations of current LLM detection methods~\cite{surveyLLMDetection}. Importantly, detection and authorship verification share methodological ground: both seek to quantify stylistic distinctiveness, but differ in whether the boundary of interest is between humans or between humans and machines.

With the advent of LLMs, a new body of work has begun to ask whether LLMs themselves can perform authorship tasks or reliably imitate writing styles. Studies have shown that prompting alone can guide models toward reproducing an individual’s stylistic cues~\cite{chen2024usingpromptsguidelarge,reynolds2021prompt}, though performance varies sharply with prompt design. Hung et al.~\cite{hung2023who} demonstrated that prompting LLMs can yield competitive results for authorship verification, while Hu et al.~\cite{InstructAV} proposed instruction fine-tuning to improve robustness further. Huang et al.~\cite{huang2024authorship} investigated whether LLMs can act as verifiers of authorship, and Scius-Bertrand et al.~\cite{scius2024zero} extended prompt-based evaluations to document classification, highlighting the general methodological relevance of zero- and few-shot prompting strategies. Beyond individual case studies, Bevendorff et al.~\cite{authorshipVerificationLLM} argued that LLM detection itself may be best framed as an authorship verification problem rather than as pure attribution, unifying two previously separate lines of research.

Taken together, prior work has advanced authorship modeling, machine-generated text detection, and LLM prompting strategies, but each in isolation. What remains missing is a systematic and reproducible framework that directly measures how closely LLM outputs match a target author’s style and how this fidelity interacts with detectability. Our work addresses this gap by combining stylometric verification techniques with LLM evaluation under controlled prompting conditions.

\begin{figure}[htp]
 \centering
 \includegraphics[width=3.2in, height= 3in]{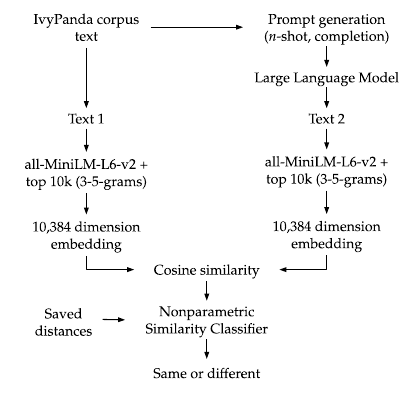}
 \caption{Framework for evaluating large language models’ ability to mimic human writing style. Human essays from IvyPanda are paired with LLM-generated counterparts under different prompting strategies. The core authorship verification pipeline embeds both texts with all-MiniLM-L6-v2 and TF–IDF features over the top 10k character 3–5 grams (10,384 dimensions). Cosine similarity between embeddings is compared against empirical distributions, and a nonparametric classifier decides whether the texts share the same author.}
 \label{fig:framework}
\end{figure}

\section{Materials and methods}
\label{MaterialsMethods}

\subsection{Protocol overview}

As summarized in Fig.~\ref{fig:framework}, our framework is a training-free authorship verification pipeline. 
The method integrates two complementary feature streams: shallow stylistic cues from TF--IDF character $n$-grams and dense contextual embeddings from a transformer encoder. 
Pairwise texts are projected into these representations and compared using cosine distance, a metric chosen for its robustness across text length regimes. 
The resulting distances are not passed into a parameterized classifier, but rather stored in empirical distributions of same-author and different-author pairs. 
Classification is then performed by non-parametric comparison against these distributions, thereby avoiding the need for hand-tuned thresholds or model training. 

Beyond the core verifier, the pipeline is extended to evaluate the capacity of large language models (LLMs) to imitate authorial style. 
We test style fidelity under four prompting conditions: zero-shot, one-shot, few-shot, and text-completion. 
Finally, we assess whether such generated texts remain statistically distinguishable from human writing by applying a perplexity-based detector. 
Together, these components form a complete experimental protocol for style-based authorship verification and imitation analysis.

\subsection{Datasets and preprocessing}
\label{subsec:data}

In Fig.~\ref{fig:framework}, the pipeline begins with textual input, which is drawn from two complementary corpora.  

\underline{Corpora:}  
For same-domain evaluation, we utilize the IvyPanda Essay corpus~\cite{ivypanda2024essays}, which contains approximately 128,000 academic essays. 
For cross-domain evaluation, we utilize the EssayForum dataset~\cite{Nidhircrossdomain}, which comprises conversational-style essays. This combination enables us to test both domain-consistent verification (academic--academic) and cross-domain robustness (academic--conversational).  

\underline{Cleaning pipeline:}  
Before entering the feature extraction modules of Fig.~\ref{fig:framework}, all texts are subjected to a deterministic filtering process designed to remove noise and artifacts. 
The criteria are: length greater than $500$ words; fewer than 10\% numeric characters; fewer than $5\%$ misspelled tokens flagged by Pyspellchecker~\cite{pyspellchecker} after lowercasing and punctuation stripping; no single token type exceeding $10\%$ frequency; fewer than $5\%$ non-punctuation symbols (e.g., \texttt{*}, \texttt{\%}, \texttt{\$}); and removal of paratextual content such as headers, footers, page numbers, and bibliographies.  
After filtering, the IvyPanda set yielded $94,942$ essays (mean length $1,561$ words), split $60/40$ into construction and evaluation partitions. The EssayForum set, after filtering, contributed $7,711$ items.  

\underline{Pair construction and materialization:}  
To generate pairs as depicted in Fig.~\ref{fig:framework} ("Text 1" and "Text 2" inputs), each essay was segmented into two non-adjacent $500$-word blocks (first and last). 
This segmentation reduces topical adjacency and encourages reliance on stylistic cues. 
All segments were encoded once into embeddings and stored, preventing repeated encoder calls during experiments. Balanced pair sets were then created: $100,000$ construction pairs ($50k$ positive, $50k$ negative), $50,000$ same-domain evaluation pairs ($25k/25k$), and $10,000$ cross-domain pairs ($5k/5k$).  

As shown later in Fig.~\ref{fig:cos_vs_eu}, performance remains stable across text lengths from $200$ to over $1000$ words, confirming that the preprocessing pipeline preserves stylistic signals while discarding irrelevant noise.  

\subsection{Feature representations and distance}
\label{subsec:features}

Once the corpora are preprocessed and segmented, Fig.~\ref{fig:framework} illustrates that each input passes through two complementary feature extraction branches: one based on surface-level $n$-grams and the other on contextual embeddings. Their outputs are then compared within the distance module.  

\underline{Character $n$-grams (TF--IDF)}
The left-hand feature branch in Fig.~\ref{fig:framework} captures shallow stylistic cues using character $n$-grams, a feature set consistently successful in PAN authorship competitions~\cite{sapkota2015not, bevendorff2020pan20}.  
From the entire training corpus, we extract the top $10^4$ character $3–5$ grams and represent each document as a sparse TF--IDF vector $\mathbf{v}_d \in \mathbb{R}^{10^4}$:  

\[
\text{tfidf}(g,d) = \frac{f(g,d)}{|d|} \cdot 
\log \!\left(\frac{N}{|\{d' \in \mathcal{D} : g \in d'\}|}\right),
\]  

where $f(g,d)$ is the frequency of $n$-gram $g$ in document $d$, $|d|$ is the total character count, and $N$ is the corpus size.  
This representation encodes frequency-based regularities such as punctuation patterns, repeated substrings, and orthographic habits, providing a surface-level lens on style.  

\underline{Transformer embeddings}  
In parallel, the right-hand branch in Fig.~\ref{fig:framework} derives dense contextual representations using the transformer encoder \texttt{all-MiniLM-L6-v2}~\cite{sentencetransformers, wang2020minilm}.  
Given a sequence of tokens $(w_1,\ldots,w_T)$, the model produces hidden states $\mathbf{h}_i \in \mathbb{R}^{384}$, which are aggregated by mean pooling:  

\[
\mathbf{e}_t = \frac{1}{T}\sum_{i=1}^T \mathbf{h}_i.
\]  

These embeddings implicitly encode authorial markers such as syntax preferences, lexical rhythm, and functional word usage~\cite{fabien2020bertaa}.  
Together with TF--IDF vectors, they provide complementary views: one explicit and frequency-based, the other implicit and distributional.  

\underline{Cosine vs.\ Euclidean distance} 
As illustrated in the central “Distance Computation” module of Fig.~\ref{fig:framework}, the two feature representations are compared using a similarity function.  
We tested cosine distance.  

\[
d_{\text{cos}}(\mathbf{x},\mathbf{y}) = 1 - \frac{\mathbf{x}\cdot \mathbf{y}}{\|\mathbf{x}\|\|\mathbf{y}\|},
\]  

against Euclidean distance  

\[
d_{\text{eucl}}(\mathbf{x},\mathbf{y}) = \|\mathbf{x} - \mathbf{y}\|_2.
\]  

\begin{figure}
\centering
\includegraphics[width=3.1in, height=1.72in]{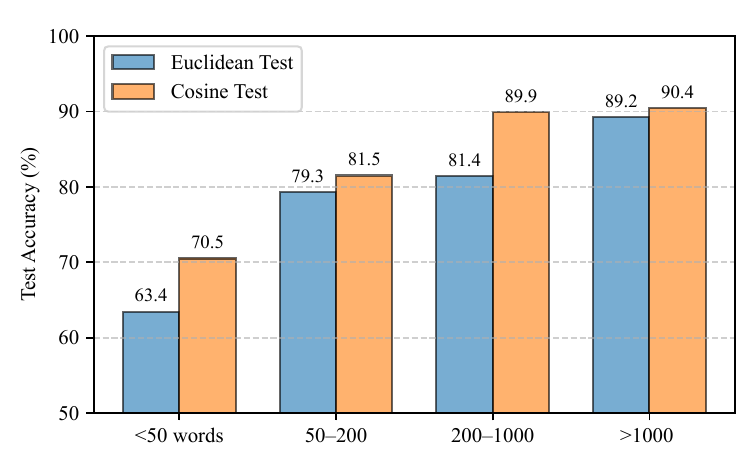}
\caption{Cosine similarity consistently outperforms Euclidean distance across all text lengths ($50$ to $1000+$ words), leading to its adoption for subsequent experiments.}
\label{fig:cos_vs_eu}
\end{figure}

Experiments across five length regimes ($50$–$1000+$ words) demonstrated that cosine similarity consistently yields sharper separation between same- and different-author pairs (Fig.~\ref{fig:cos_vs_eu}). This robustness arises because the cosine function normalizes for vector magnitude, thereby mitigating biases from text length and frequency scaling. Accordingly, cosine distance was adopted as the standard metric throughout the framework.  

\subsection{Training-free distribution verifier}
\label{subsec:verifier}

After pairwise distances are computed, Fig.~\ref{fig:framework} shows that they feed into the distribution-based verification module.  
Instead of training a classifier with learnable parameters, the verifier stores empirical distance distributions for same-author and different-author pairs, enabling classification without thresholds or gradient updates.  

\subsubsection{\underline{Construction }}  
During construction, distances are computed for all labeled training pairs.  
Formally, for $(t_i, t_j) \in \mathcal{P}^+$ (same-author) and $(t_i, t_j) \in \mathcal{P}^-$ (different-author), we calculate:  

\[
d_{ij} = d_{\text{cos}}(\mathbf{e}_{t_i}, \mathbf{e}_{t_j}).
\]  

These values populate two empirical distributions:  

\[
\begin{aligned}
\mathcal{D}^+ &= \{ d_{ij} : (t_i, t_j) \in \mathcal{P}^+ \}, \\
\mathcal{D}^- &= \{ d_{ij} : (t_i, t_j) \in \mathcal{P}^- \}.
\end{aligned}
\]

As indicated by the \textit{saved distances} block in Fig.~\ref{fig:framework}, this step produces reference baselines against which all test pairs are later compared.  
No weights, thresholds, or parameter updates are learned.  

\subsubsection{\underline{Decision rule and confidence}}  
During inference, a test pair $(t_a, t_b)$ yields distance $d^*$.  
As shown in the “Nonparametric Similarity Classifier” block of Fig.~\ref{fig:framework}, the verifier evaluates:  

\[
S = \Pr_{\delta \sim \mathcal{D}^+}[\delta > d^*], \qquad 
D = \Pr_{\delta \sim \mathcal{D}^-}[\delta < d^*].
\]  

The rule is simple: if $S > D$, classify as same-author; otherwise, different-author.  
A confidence score quantifies the separation:  

\[
\text{Confidence} = \frac{|S - D|}{\max(S,D)}.
\]  

This measure reflects how decisively the test distance lies within one distribution rather than the other.  

\subsubsection{\underline{Relation to nonparametrics and ablation}}  
This design generalizes nonparametric classifiers such as $k$-NN~\cite{cover1967nearest}.  
However, instead of local neighborhood voting, it leverages global distance distributions, eliminating hyperparameters such as $k$ or threshold tuning.  
Ablation experiments confirm the value of this approach: a handcrafted 16-feature Siamese network built on spaCy~\cite{spacy} achieved only $57\%$ accuracy, whereas the distributional verifier—driven by TF--IDF $n$-grams and transformer embeddings (Fig.~\ref{fig:framework}, dual feature inputs)—achieved substantially higher performance.  

\subsection{LLM Style imitation setup}
\label{subsec:llmsetup}

Beyond human-authored corpora, Fig.~\ref{fig:framework} illustrates a second pathway where large language models (LLMs) generate candidate texts, which are then fed into the same feature-extraction and verification pipeline. This component evaluates whether modern LLMs can replicate an individual author’s style closely enough to deceive the distribution-based verifier.  

\subsubsection{\underline{Models and common prompt preamble}}  
We selected five LLMs from three different families: Llama-$3.3$-$70$B-Instruct~\cite{llama3.3-hf}, Llama-$4$-Scout-$17$B-$16$E-Instruct~\cite{llama4-scout-hf}, Mixtral $8$x$7$B~\cite{jiang2024mixtralexperts}, Qwen $2.5$ $14$B Instruct~\cite{qwen2025qwen25technicalreport}, and Qwen $2.5$ $32$B Instruct~\cite{qwen2025qwen25technicalreport}.  
These models differ in scale, architecture, and generation, allowing us to test a diverse set of LLMs and determine universal trends.
To standardize outputs, every experiment included the same preamble in the system prompt:  

\begin{quote}
\textit{``Output only the requested content. No prefaces, disclaimers, or explanations.''}
\end{quote}

This prevented models from adding meta-text and ensured compatibility with the pipeline in Fig.~\ref{fig:framework}, where synthetic outputs enter the same feature branches (TF--IDF and transformer embeddings) as human texts.  

\subsubsection{\underline{Prompting conditions}}  
We tested four prompting conditions, each designed to separate style imitation from topical overlap:  

\begin{itemize}
    \item \textit{Zero-shot:} the model receives only a statistical profile of the target author (syntax counts, punctuation ratios, common bigrams). It generates a $300-500$ word essay on any topic, relying solely on style-level cues.  
    \item \textit{One-shot:} the model is given a single longest paragraph from the target author as an anchor. It is explicitly instructed to write on a different topic, ensuring that imitation cannot be achieved through simple topical continuation.  
    \item \textit{Few-shot:} the two longest paragraphs are supplied as anchors. This richer context allows models to form a clearer stylistic template.  
    \item \textit{Completion:} Each human essay is split in half by length. The model is asked to continue the first half with a passage of similar length, without repetition or explicit reference to the original passage. This scenario reflects realistic use cases for co-writing or auto-completion.  
\end{itemize}

For each condition, $1,000$ held-out authors were sampled per model. The generated texts were stored and subsequently fed through the same verification pipeline (Fig.~\ref{fig:framework}, "Text $1$" and "Text $2$" inputs).  
This design ensured that comparisons between human and machine-authored essays were performed under the same feature and decision rules described in \S\ref{subsec:features} and \S\ref{subsec:verifier}.  

\subsection{Evaluation metrics}
\label{subsec:metrics}
Let $\{(x_i,y_i)\}_{i=1}^n$ denote the test set, where $x_i$ is a pair of texts and $y_i \in \{0,1\}$ is the ground-truth label ($1 =$ same-author, $0 =$ different-author).  
The verifier outputs a predicted label $\hat{y}_i$ and a confidence score $s_i \in [0,1]$.

\underline{Accuracy:}  
Accuracy is the proportion of test pairs for which the predicted label matches the ground truth:
\[
\text{Acc} = \frac{1}{n}\sum_{i=1}^n \mathbb{1}[\hat{y}_i = y_i],
\]
where $\mathbb{1}[\cdot]$ is the indicator function, equal to $1$ if the condition holds and $0$ otherwise.  
This corresponds to the fraction of correctly identified pairs, whether same-author (true positives) or different-author (true negatives).

\underline{ROC AUC:}  
Let $\mathcal{S}^+ = \{s_i : y_i=1\}$ be the confidence scores for same-author pairs and $\mathcal{S}^- = \{s_i : y_i=0\}$ the scores for different-author pairs.  
For a decision threshold $\tau \in [0,1]$, the true positive rate (TPR) and false positive rate (FPR) are
\begin{align*}
\text{TPR}(\tau) &= \frac{|\{s \in \mathcal{S}^+ : s \geq \tau\}|}{|\mathcal{S}^+|}, \\
\text{FPR}(\tau) &= \frac{|\{s \in \mathcal{S}^- : s \geq \tau\}|}{|\mathcal{S}^-|}.
\end{align*}

The ROC curve plots TPR against FPR as $\tau$ varies.  
The area under the curve (AUC) is
\[
\text{AUC} = \Pr(s^+ > s^-), \quad s^+ \sim \mathcal{S}^+, \; s^- \sim \mathcal{S}^-,
\]
which is the probability that a randomly chosen same-author pair receives a higher confidence score than a randomly chosen different-author pair. In this setting, AUC quantifies how well the verifier distinguishes genuine stylistic matches from non-matches, including LLM-generated imitations.

\underline{Confusion matrix:}  
The verifier’s predictions can be summarized as follows:  

\[
\begin{array}{c|c|c}
 & \text{Pred.\ Same} & \text{Pred.\ Diff.} \\
\hline
\text{Actual Same} & \text{TP} & \text{FN} \\
\text{Actual Diff.} & \text{FP} & \text{TN}
\end{array}
\]

Formally and in context:  

\[
\text{TP} = |\{i : y_i=1, \hat{y}_i=1\}| \]
(same-author pairs correctly identified as same; e.g., two human essays or a human essay with its LLM continuation),

\[
\text{TN} = |\{i : y_i=0, \hat{y}_i=0\}| \]

(different-author pairs correctly identified as different; e.g., two humans or a human and unrelated LLM output),

\[
\text{FP} = |\{i : y_i=0, \hat{y}_i=1\}| \]
(different-author pairs misclassified as the same; overestimation of stylistic similarity, e.g., an LLM imitation mistaken for the author),

\[
\text{FN} = |\{i : y_i=1, \hat{y}_i=0\}| \]
(same-author pairs misclassified as different; underestimation of genuine stylistic consistency).

\underline{McNemar’s test:}  
This test evaluates whether two systems (e.g., TF--IDF vs.\ transformer embeddings) show a significant difference in verification accuracy.  

Formally, let $A(x_i)$ and $B(x_i)$ denote the predictions of two systems on input pair $x_i$ with ground truth $y_i$:  

\[
n_{01} = |\{i : A(x_i) \neq y_i, \; B(x_i) = y_i\}| \]
(pairs misclassified by $A$ but correctly classified by $B$),

\[
n_{10} = |\{i : A(x_i) = y_i, \; B(x_i) \neq y_i\}| \]
(pairs correctly classified by $A$ but misclassified by $B$).

The McNemar statistic is  
\[
\chi^2 = \frac{(|n_{01} - n_{10}| - 1)^2}{n_{01} + n_{10}},
\]  

which follows a $\chi^2$ distribution with one degree of freedom under the null hypothesis of no performance difference.  
A significant result ($p < 0.05$) indicates that one system captures authorial style more faithfully than the other.  

\underline{Verifier confidence:}  
For a test pair with distance $d^*$, let $\mathcal{D}^+$ and $\mathcal{D}^-$ denote the empirical distributions of same-author and different-author distances (\S\ref{subsec:verifier}). We compute \[
S(d^*) = \Pr_{\delta \sim \mathcal{D}^+}[\delta > d^*] \]
(prob. that $d^*$ is closer to the same-author distribution),
 
 \[
D(d^*) = \Pr_{\delta \sim \mathcal{D}^-}[\delta < d^*] \]
(prob. that $d^*$ is closer to the different-author distrib).

The decision rule is   \[
\hat{y} = 
\begin{cases} 
1 & \text{if } S(d^*) > D(d^*) \quad \text{(predicted same-author)} \\ 
0 & \text{otherwise} \quad \text{(predicted different-author)} 
\end{cases}
\]

with confidence score \[
\text{Conf}(d^*) = \frac{|S(d^*) - D(d^*)|}{\max(S(d^*), D(d^*))}.
\]

This scalar quantifies how decisively the test distance aligns with one distribution. High values indicate a clear separation (the pair is well inside one distribution), while low values reflect overlap or ambiguity. In practice, this measures how strongly the verifier supports its decision about whether an LLM output matches or deviates from a human author’s style.  

\section{Results}
\label{Results}
 
\subsection{Authorship verification results}

The distribution-based verifier achieved consistently strong results across both same-domain (IvyPanda) and cross-domain (EssayForum) evaluations. As summarized in Table~\ref{tab:verification-results}, same-domain accuracy reached $97.49\%$ with a ROC AUC of $0.997$, indicating near-perfect separation of same- and different-author pairs. The corresponding F1 score of $0.975$ reflects balanced precision and recall, while the mean confidence of $97.1\%$ ($\pm 11$) confirms that most distances fell deep inside the correct distribution.  

Cross-domain testing naturally reduced performance due to a distributional shift, resulting in a decrease in accuracy to $94.48\%$ and F1 to $0.870$, while the mean confidence dropped to $88.4\%$ ($\pm 4\%$). Nevertheless, ROC AUC remained high at $0.981$, demonstrating that the verifier maintained strong discriminative power even when generalizing across domains.  

\begin{table}[ht]
\centering
\scriptsize
\caption{Authorship verification performance across same- and cross-domain evaluation via different metrics and their interpretations.}
\label{tab:verification-results}
\begin{tabular}{lrrl}
\hline
\textbf{Metric} & \textbf{Same} & \textbf{Cross} & \textbf{Interpretation} \\
\hline
Accuracy & 97.49 & 94.48 & Overall correctness \\
ROC AUC & 0.997 & 0.981 & Discriminability \\
F1 Score & 0.975 & 0.870 & Precision--recall balance \\
Confidence & 97.1\% $\pm$ 11 & 88.4\% $\pm$ 4 & Avg.\ separation strength \\
McNemar $\chi^2$ & 
\shortstack{627.0 \\ ($p<0.05$)} & 
\shortstack{1349.0 \\ ($p<0.05$)} & 
Significant \\
\hline
\end{tabular}
\end{table}

\begin{figure}[ht]
    \centering
    \begin{subfigure}{.99\linewidth}
        \centering
        \includegraphics[width=3.1in, height=2.1in]{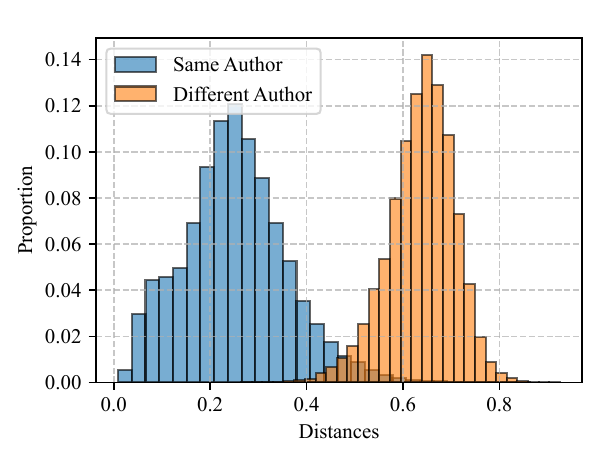}
        \caption{Histogram same vs. different author distances.}
        \label{fig:perplexity_hist}
    \end{subfigure}
    
    \vspace{1em} 
    
    \begin{subfigure}{0.99\linewidth}
        \centering
        \includegraphics[width=3.1in, height=2.1in]{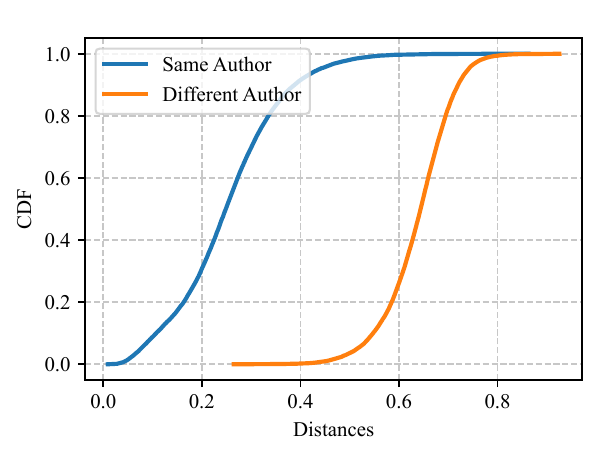}
        \caption{CDF of same vs. different author distances.}
        \label{fig:perplexity_cdf}
    \end{subfigure}
    
    \caption{Distance distributions used for authorship verification (cosine distance between embedding pairs). (a) Normalized histograms show strong separation: same-author pairs concentrate at lower distances (around $0.25$) and different-author pairs at higher distances (around $0.65$), with limited overlap. (b) Empirical CDFs, independent of binning, make the gap explicit: thresholds in the middle range yield a high true-positive rate for same-author pairs, accompanied by a low false-positive rate for different-author pairs. These distributions motivate the paper’s distribution-based, threshold-free decision rule.}
    \label{fig:hist-cdf-distances}
\end{figure}

The distance distributions in Fig.~\ref{fig:hist-cdf-distances} explain these results. In the same-domain setting, same-author pairs cluster tightly around low cosine distances (around $0.25$), while different-author pairs concentrate near higher distances (around $0.65$). The small overlap region accounts for the symmetric errors observed in the confusion matrix (Table~\ref{tab:confusion-matrix-both}, FN = $627$, FP = $628$).  

In contrast, the cross-domain evaluation reveals a marked asymmetry: $1,349$ false negatives versus only $113$ false positives. This pattern suggests that stylistic drift in conversational essays draws genuine same-author pairs closer to the distribution of different authors, thereby reducing recall while preserving precision.  

\begin{table}[ht]
\centering
\scriptsize
\caption{Confusion matrices for same- and cross-domain evaluation. Percentages are relative to row totals.}
\label{tab:confusion-matrix-both}
\begin{tabular}{llrr}
\hline
 & & \textbf{Pred.\ Same} & \textbf{Pred.\ Diff.} \\
\hline
\multirow{2}{*}{\textbf{Same}} 
 & Actual Same     & 24,373 (97.5\%) & 627 (2.5\%) \\
 & Actual Different & 628 (2.5\%)    & 24,372 (97.5\%) \\
\hline
\multirow{2}{*}{\textbf{Cross}} 
 & Actual Same     & 3651 (73.0\%)  & 1349 (27.0\%) \\
 & Actual Different & 113 (2.3\%)   & 4887 (97.7\%) \\
\hline
\end{tabular}
\end{table}

\subsection{LLM Style imitation: full-text generation}

Prompting strategy emerged as the dominant factor in style imitation. As shown in Table~\ref{tab:llm-style-results}, all models failed in the zero-shot condition (accuracy below $7\%$) with a few texts fooling the verification model. Additionally, the verifier reported high confidence ($>95\%$) for all its predictions, cementing that zero-shot prompts are indeed incapable of style mimicry. This suggests that the statistical style summaries provided in prompts were not effective anchors for imitation. One-shot results in Table~\ref{tab:llm-style-results} improve drastically, but intra-prompting-strategy accuracies varied wildly ($67.6\%$ to $94.7\%$). This sparsity in accuracy shows no clear correlation to model generation architecture or alignment strategy, indicating that prompting strategy, rather than underlying model design, dominates style fidelity outcomes. This is concurred by the few-shot results, which, while they vary slightly between models, show a strong upward trend from one-shot results.

\begin{table}[ht]
\centering
\scriptsize
\caption{Authorship imitation performance across prompting strategies for five diverse models. Accuracy (\%) is reported.}
\label{tab:llm-style-results}
\begin{tabular}{l|r|r|r|r}
\hline
\textbf{Model} & \textbf{Zero} & \textbf{One} & \textbf{Few} & \textbf{Comp} \\
\hline
Mixtral 8x7B         & 6\%   & 74\%   & 91\%   & 100\% \\
Qwen 2.5 14B         & 5.26\% & 94.7\% & 100\%  & 99\% \\
Qwen 2.5 32B         & 3.1\%  & 89.9\% & 96.6\% & 96.9\% \\
Llama-3.3-70B        & 6.9\%  & 78.2\% & 95\%   & 99.9\% \\
Llama-4-Scout-17B    & 4.2\%  & 67.6\% & 98.8\% & 99.9\% \\
\hline
\end{tabular}
\end{table}

\subsection{LLM style imitation: text completion}
The continuation scenario produced the strongest results. As shown in Table~\ref{tab:llm-style-results}, four out of five models achieved at least $99.9\%$ accuracy with verifier confidence saturated at $100\%$. Because the first half of each essay establishes a strong stylistic manifold, both models remain within it during continuation, rendering their outputs virtually indistinguishable from the original author. This finding has direct implications for forensic analysis: once a stylistic context is provided, continuation tasks may evade detection by stylometric means.  

\begin{table}[htp]
\centering
\scriptsize
\caption{Average perplexity of LLM (i.e, Llama) outputs by prompting strategy (human baseline: $29.5$).}
\label{tab:perplexity-results}
\begin{tabular}{lrrl}
\hline
\textbf{Strategy} & \textbf{3.3-70B} & \textbf{4-Scout-17B} & \textbf{Interpretation} \\
\hline
Zero-shot & 8.90 & 10.46 & Most predictable \\
One-shot  & 17.45 & 19.87 & Higher variability \\
Few-shot  & 16.43 & 15.16 & Model-dependent shift\\
\hline
\end{tabular}
\end{table}

\begin{figure}[ht]
    \centering
    
    \begin{subfigure}{0.99\linewidth}
        \centering
        \includegraphics[width=3.1in, height=2.1in]{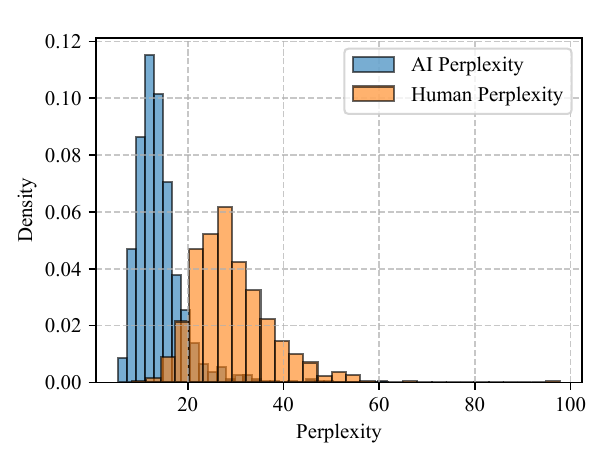}
        \caption{Histogram comparison of AI vs. Human perplexity.}
        \label{fig:perplexity_hist}
    \end{subfigure}
    
    \vspace{1em} 
    
    \begin{subfigure}{0.99\linewidth}
        \centering
        \includegraphics[width=3.1in, height=2.1in]{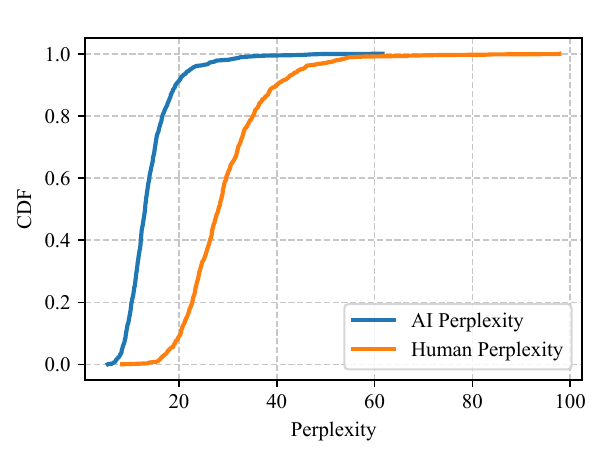}
        \caption{CDF comparison of AI vs. Human perplexity.}
        \label{fig:perplexity_cdf}
    \end{subfigure}
    
    \caption{Perplexity distributions for AI- and human-written essays computed with GPT-2. (a) The histogram (density scale, common bin edges) shows that AI texts are concentrated near $15$ ($\mu=15.2$), while human texts center around $30$ ($\mu = 29.5$). (b) Empirical CDFs make the separation independent of binning: at perplexity $\leq 20$, about $90\%$ of AI texts and about $10-15\%$ of human texts fall below the threshold; at $\leq 30$, about $99\%$ of AI and about $55-60\%$ of human texts fall below. Lower perplexity indicates greater predictability; therefore, both views suggest that AI-generated outputs remain more predictable than human-written texts, even when the style is similar.}
    \label{fig:perplexity_combined}
\end{figure}

\subsection{AI detectability: Perplexity analysis}
Despite their stylistic fidelity, LLM outputs remain substantially more predictable than human writing. As shown in Figure~\ref{fig:perplexity_combined}, IvyPanda essays average a perplexity of \textit{$29.5$}, while generated texts average only \textit{$16.07$}. At thresholds $\leq 20$, nearly $90\%$ of AI outputs fall below compared to just $15\%$ of human essays, indicating a clear detectability gap.  

Detectability shows no clear correlation with prompting strategy or style fidelity. Other than four out of five zero-shot averages having noticeably lower perplexity scores, the remainder are scattered seemingly at random. So while LLM text in general is algorithmically predictable, this doesn't appear to correlate to style-mimicry-quality.

\section{Discussion}
\label{Discussion}

The experiments provide a comprehensive view of how authorship verification and style imitation interact under controlled conditions. Several findings stand out.

First, the distribution-based verifier demonstrated both effectiveness and efficiency. On IvyPanda, it reached $97.5\%$ accuracy, ROC AUC of $0.997$, and an F1 of $0.975$, with confidence concentrated around $97.1\%$ ($\pm$11). These values indicate that the empirical distance distributions accurately capture authorial consistency with minimal overlap. The confusion matrix confirmed this balance, with nearly identical false positives ($628$) and false negatives ($627$), suggesting that errors stemmed primarily from the natural overlap in writing styles rather than systematic bias. Cross-domain evaluation on EssayForum, although reduced to $94.5\%$ accuracy and $0.870$ F1, still achieved an AUC of $0.981$, underscoring its robustness. Here, the asymmetry in errors—$1349$ false negatives but only $113$ false positives—highlights a key challenge: conversational writing by the same author is often judged to be more distant from their academic writing than essays from other authors within the same domain. This pattern suggests that robustness to genre and register remains a limiting factor in real-world deployment.

Second, the imitation experiments clarify the role of prompting. In zero-shot settings, accuracy dropped below $7\%$ for all five models, despite confidence exceeding $90\%$, implying that stylistic profiles alone were insufficient anchors. Accuracy improved in one-shot prompts ($67.6–94.7\%$), but with significant variance, reflecting unstable style reproduction. Few-shot prompting showed further improvements from one-shot prompts from all models. Completion prompts proved strongest, with four-fifths of the models achieving $99.9\%$ accuracy and verifier confidence saturated at $100\%$. This indicates that once a human-authored prefix establishes the stylistic manifold, models can maintain it with remarkable fidelity. Forensic implications are immediate: completion scenarios blur the line between co-writing and imitation, making detection far more difficult.

Third, perplexity analysis shows that fidelity and detectability are separable. Human essays averaged $29.5$ perplexity, compared to $16.07$ for LLM outputs. At thresholds $\leq 20$, about $90\%$ of generated texts fell below, versus only $15\%$ of human essays. Prompting strategy and stylistic fidelity appear not to influence the algorithmic predictability. Ultimately, even as models achieve near-perfect imitation, their outputs remain algorithmically regular. 

Taken together, these results highlight both the promise and risk of current LLMs. On the one hand, training-free verification is reliable, scalable, and interpretable, with performance rivaling supervised baselines at a fraction of the cost. On the other hand, LLMs equipped with exemplar-based prompts can achieve style imitation accuracy indistinguishable from human authorship, raising challenges for academic integrity and forensic attribution. Importantly, statistical predictability persists even in high-fidelity imitation, underscoring that authorship verification and AI detection should be treated as complementary rather than redundant tasks. Extending this analysis to broader genres, multilingual settings, and adversarial prompt designs represents the next frontier in understanding the stylistic capabilities and limitations of LLMs.

\section{Limitations and Future Work}
\label{Limitations}
This work was evaluated on English academic and conversational essays, leaving open questions about generality across creative, technical, and multilingual domains. Broader corpora are needed to test whether the observed results hold across diverse registers of authorial style.  

Although the verifier is efficient and training-free, it remains retrospective and may be challenged by adversarial tactics such as paraphrasing or prompt manipulation. Likewise, perplexity-based detection, here anchored on GPT-$2$, offers only a partial view; newer detectors or multi-metric approaches that capture syntactic or discourse-level irregularities may reveal different boundaries between human and machine text.  

Future work should extend evaluation to a wider range of LLM architectures and prompting methods, and explore proactive safeguards such as watermarking or identity-conditioned generation. Together, these directions will help establish more robust and preventive frameworks for authorship verification in the era of large-scale generative models.
 
\section{Conclusion}
\label{Conclusion}
We introduced a training-free, distribution-based verifier that fuses TF–IDF character $n$-grams with transformer embeddings to quantify stylistic similarity without thresholds or supervised fitting, achieving reproducible, state-of-the-art performance—$97.49\%$ accuracy (AUC $0.997$, F1 $0.975$) in-domain and $94.48\%$ (AUC $0.981$, F1 $0.870$) cross-domain—while reducing construction time by $91.8\%$ and memory by $59\%$ relative to parameterized baselines; error patterns were interpretable (symmetric FP/FN in-domain; FN-heavy under genre drift), aligning with observed distance distributions. Leveraging this verifier, we showed that prompting strategy, not model size, primarily governs style imitation: zero-shot failed, one-shot improved substantially, few-shot reached near-perfect alignment, and 80\% of models in text completion attained $99.9\%$ agreement once a human prefix established the stylistic manifold. Despite this fidelity, generated text remained more predictable than human writing (perplexity $16.07$ vs.\ $29.5$; at threshold $20$, $\sim90\%$ of AI vs.\ $\sim15\%$ of human texts fell below), demonstrating that \emph{fidelity and detectability are separable}. These results provide a scalable, interpretable basis for style-aware evaluation, clarify the centrality of exemplar-based prompting for reliable imitation, and motivate dual-track safeguards that pair authorship verification with predictability-based detection in identity-conditioned generation.

\balance
\bibliography{references-long}

\end{document}